\documentclass[conference,letterpaper]{IEEEtran}
\IEEEoverridecommandlockouts
\usepackage{subfigure,color}
\usepackage{algorithm}
\usepackage{algpseudocode}
\usepackage{amssymb}
\usepackage{amsmath}
\usepackage{lscape}
\usepackage{multirow}
\usepackage{url}

\usepackage{cite}
\usepackage{amsmath,amssymb,amsfonts}
\usepackage{graphicx}
\usepackage{textcomp}
\usepackage{xcolor}
\def\BibTeX{{\rm B\kern-.05em{\sc i\kern-.025em b}\kern-.08em
    T\kern-.1667em\lower.7ex\hbox{E}\kern-.125emX}}
\begin{document}

\title{Deep Reinforcement Learning for Chatbots Using Clustered Actions and Human-Likeness Rewards
\thanks{Work carried out while the first author was visiting Samsung Research.}
}

\author{\IEEEauthorblockN{Heriberto Cuay\'ahuitl}
\IEEEauthorblockA{\textit{School of Computer Science} \\
\textit{University of Lincoln}\\
Lincoln, United Kingdom \\
HCuayahuitl@lincoln.ac.uk}
\and
\IEEEauthorblockN{Donghyeon Lee}
\IEEEauthorblockA{\textit{Artificial Intelligence Research Group} \\\textit{Samsung Electronics}\\
Seoul, South Korea \\
dh.semko.lee@samsung.com}
\and
\IEEEauthorblockN{Seonghan Ryu}
\IEEEauthorblockA{\textit{Artificial Intelligence Research Group} \\\textit{Samsung Electronics}\\
Seoul, South Korea \\
seonghan.ryu@samsung.com}
\and
\IEEEauthorblockN{Sungja Choi}
\IEEEauthorblockA{\textit{Artificial Intelligence Research Group} \\\textit{Samsung Electronics}\\
Seoul, South Korea \\
sungja.choi@samsung.com}
\and
\IEEEauthorblockN{Inchul Hwang}
\IEEEauthorblockA{\textit{Artificial Intelligence Research Group} \\\textit{Samsung Electronics}\\
Seoul, South Korea \\
inc.hwang@samsung.com}
\and
\IEEEauthorblockN{Jihie Kim}
\IEEEauthorblockA{\textit{Artificial Intelligence Research Group} \\\textit{Samsung Electronics}\\
Seoul, South Korea \\
jihie.kim@samsung.com}
}

\maketitle

\begin{abstract}
Training chatbots using the reinforcement learning paradigm is challenging due to high-dimensional states, infinite action spaces and the difficulty in specifying the reward function. We address such problems using clustered actions instead of infinite actions, and a simple but promising reward function based on human-likeness scores derived from human-human dialogue data. We train Deep Reinforcement Learning (DRL) agents using chitchat data in raw text---without any manual annotations. Experimental results using different splits of training data report the following. First, that our agents learn reasonable policies in the environments they get familiarised with, but their performance drops substantially when they are exposed to a test set of unseen dialogues. Second, that the choice of sentence embedding size between 100 and 300 dimensions is not significantly different on test data. Third, that our proposed human-likeness rewards are reasonable for training chatbots as long as they use lengthy dialogue histories of $\geq$10 sentences.
\end{abstract}


\section{Introduction}
\label{intro}
What happens in the minds of humans during chatty interactions containing sentences that are not only coherent but also engaging? While not all chatty human dialogues are engaging, they are arguably coherent \cite{GroszS86}. They also exhibit large vocabularies---according to the language in focus---because conversations can address any topic that comes to the minds of the partner conversants. In addition, each contribution by a partner conversant may exhibit multiple sentences instead of one such as greeting+question or acknowledgement+statement+question. Furthermore, the topics raised in the conversation may go back and forth without losing coherence. This is a big challenge for data-driven chatbots.

\begin{figure}[th!]
  \centerline{\includegraphics[width=8.8cm]{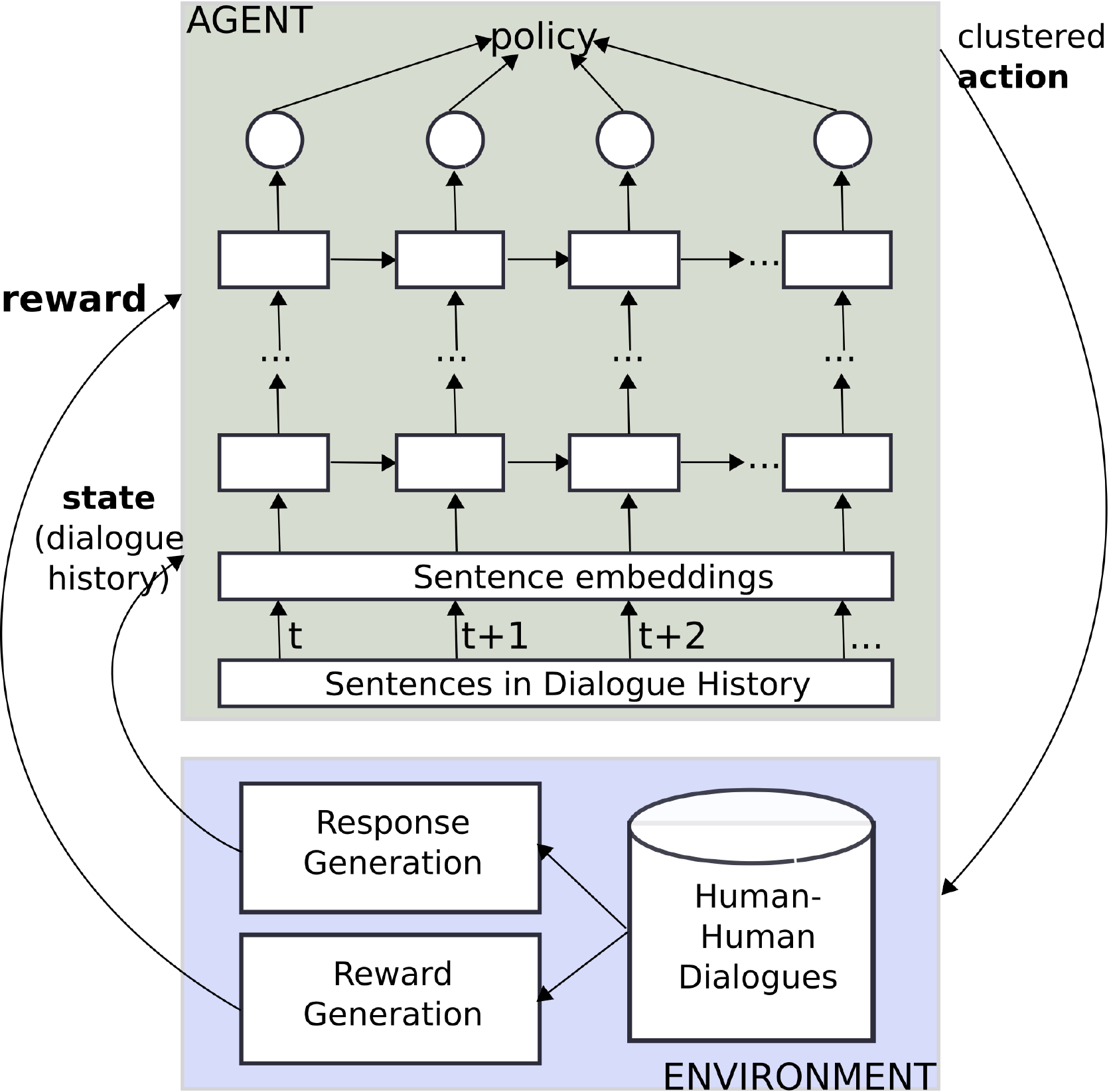}}
%
\caption{High-level architecture of the proposed deep reinforcement learning approach for chatbots---see text for details}
\label{agent-arch}
\end{figure}

We present a novel approach based on the reinforcement learning \cite{SuttonB2018}, unsupervised learning \cite{HastieTF09} and deep learning \cite{LeCunBH15} paradigms. Our learning scenario is as follows: given a data set of human-human dialogues in raw text (without any manually provided labels), a Deep Reinforcement Learning (DRL) agent takes the role of one of the two partner conversants in order to learn to select human-like sentences when exposed to both human-like and non-human-like sentences. In our learning scenario the agent-environment interactions consist of agent-data interactions -- there is no user simulator as in task-oriented dialogue systems. During each verbal contribution, the DRL agent (1) observes the state of the world via a deep neural network, which models a representation of all sentences raised in the conversation together with a set of candidate responses or agent actions (referred as {\it clustered actions} in our approach); (2) it then selects an action so that its word-based representation is sent to the environment; and (3) it receives an updated dialogue history and a numerical reward for having chosen each action, until a termination condition is met. This process---illustrated in Figure~\ref{agent-arch}---is carried out iteratively until the end of a dialogue for as many dialogues as necessary, i.e. until there is no further improvement in the agent's performance. 

The contributions of this paper are as follows.
\begin{itemize}
\item We propose to train chatbots using value-based deep reinforcement learning using action spaces derived from unsupervised clustering, where each action cluster is a representation of a type of meaning (greeting, question around a topic, statements around a topic, etc.).
\item We propose a simple though promising reward function. It is based on human-human dialogues and noisy dialogues for learning to rate good vs. bad dialogues. According to an analysis of dialogue reward prediction, dialogues with lengthy dialogue histories (of at least 10 sentences) report strong correlations between true and predicted rewards on test data.
\item Our experiments comparing different sentence embedding sizes (100 vs. 300) did not report statistical differences on test data. This means that similar results can be obtained more efficiently with the smaller embedding than the larger one due to less features. In other words, sentence embeddings of 100 dimensions are as good as 300 dimensions but less computationally demanding.
\item Last but not least, we found that training chatbots on multiple data splits is crucial for improved performance over training chatbots using the entire training set.
\end{itemize}

The remainder of the paper describes our proposed approach in more detail and evaluates it using a publicly available dataset of chitchat conversations. Although our learning agents indeed improve their performance over time with dialogues that they get familiarised with, their performance drops with dialogues that the agents are not familiar with. The former is promising and in favour of our proposed approach, and the latter is not, but it is a general problem faced by data-driven chatbots and an interesting avenue for future research.

\section{Related Work}
\label{litreview}

Reinforcement Learning (RL) methods are typically based on value functions or policy search \cite{SuttonB2018}, which also applies to deep RL methods. While value functions have been particularly applied to task-oriented dialogue systems \cite{CasanuevaBSURTG18,Cuayahuitl16,CuayahuitlYWC17,CuayahuitlY17,WilliamsAZ17,PengLLGCLW17}, policy search has been particularly applied to open-ended dialogue systems such as (chitchat) chatbots \cite{LiMSJRJ16,LiMSJRJ17,SerbanEtAl2018,SankarRavi2018neurips,abs-1804-02504}. This is not surprising given the fact that task-oriented dialogue systems use finite action sets, while chatbot systems use infinite action sets. So far there is a preference for policy search methods for chatbots, but it is not clear whether they should be preferred because they face problems such as local optima rather than global optima, inefficiency and high variance. It is thus that this paper explores the feasibility of value function-based methods for chatbots, which has not been explored before---at least not from the perspective of deriving the action sets automatically as attempted in this paper. 

Other closely related methods to deep RL include seq2seq models for dialogue generation \cite{VinyalsL15,SordoniGABJMNGD15,SerbanKTTZBC17,LiGBSGD16,Wang2018,ZhangEtAl2018}. These methods tend to be data-hungry because they are typically trained with millions of sentences, which imply high computational demands. 
While they can be used to address the same problem, in this paper we focus our attention on deep RL-based chatbots and leave their comparison or combination as future work. Nonetheless, these related works agree with the fact that evaluation is a difficult part and that there is a need for better evaluation metrics \cite{Yan18}. This is further supported by \cite{LiuLSNCP16}, where they found that metrics such as Bleu and Meteor amongst others do not correlate with human judgments. 

With regard to performance metrics, the reward functions used by deep RL dialogue agents are either specified manually depending on the application, or learnt from dialogue data. For example, \cite{LiMSJRJ16} conceives a reward function that rewards positively sentences that are easy to respond and coherent while penalising repetitiveness. \cite{LiMSJRJ17} uses an adversatial approach, where the discriminator is trained to score human vs. non-human sentences so that the generator can use such scores during training. \cite{SerbanEtAl2018} trains a reward function from human ratings. All these related works are neural-based, and there is no clear best reward function to use in future (chitchat) chatbots. This motivated us to propose a new metric that is easy to implement, practical due to requiring only data in raw text, and potentially promising as described below.

\section{Proposed Approach}
\label{approach}

To explain the proposed learning approach we first describe how to conceive a finite set of dialogue actions from raw text, then we describe how to assign rewards, and finally describe how to bring everything together during policy learning. 

\subsection{Clustered Actions}
\label{clusteredactions}
Actions in reinforcement learning chatbots correspond to sentences, and their size is infinite assuming all  possible combinations of words sequences in a given language. This is especially true in the case of open-ended conversations that make use of large vocabularies, as opposed to task-oriented conversations that make use of smaller (restricted) vocabularies. A {\bf clustered action} is a group of sentences sharing a similar or related meaning via {\it sentence vectors} derived from word embeddings \cite{MikolovSCCD13,PenningtonSM14}. While there are multiple ways of selecting features for clustering and also multiple clustering algorithms, the following requirements arise for chatbots: (1) unlabelled data due to human-human dialogues in raw text (this makes it difficult to evaluate the goodness of clustering features and algorithms), and (2) scalability to clustering a large set of data points (sentences in our case, which are mostly unique).

Given a set of data points $\{{\bf x}_1,\cdots,{\bf x}_n\} \forall {\bf x}_i \in \mathbb{R}^m$ and a similarity metric 
$d({\bf x}_i,{\bf x}_{i'})$, the task is to find a set of $k$ clusters with a clustering algorithm. Since in our case each data point ${\bf x}$ corresponds to a sentence within a dialogue, we represent sentences via their mean word vectors---similarly as in Deep Averaging Networks \cite{IyyerMBD15}---denoted as 
\begin{equation}\nonumber
{\bf x}_i=\frac{1}{N_i}\sum_{j=1}^{N_i} c_j,
\end{equation} 
where $c_j$ is the vector of coefficients of word $j$ and $N_i$ is the number of words in sentence $i$. For scalability purposes, we use the K-Means++ algorithm \cite{ArthurV07} with the Euclidean distance 
\begin{equation}\nonumber
d({\bf x}_i^j,{\bf x}_{i'}^j)=\sqrt{\sum_{j=1}^m({\bf x}_i^j,{\bf x}_{i'}^j)^2}
\end{equation}
with $m$ dimensions, and assume that $k$ is provided rather than automatically induced -- though other algorithms can be used with our approach. In this way, a trained clustering model assigns a cluster ID $a \in A$ to features ${\bf x}_i$, where the number of actions is equivalent to the number of clusters, i.e. $\vert A \vert = k$.

\begin{table*}[th!]
\begin{center}
\begin{tabular}{|l|l|} 
\hline
{\bf Human Sentences} & {\bf Distorted Human Sentences} \\
\hline
\hline
hello what are doing today? & hello what are doing today? \\ 
\textcolor{blue}{i'm good, i just got off work and tired, i have two jobs.}[$r$=+1] & \textcolor{red}{do your cats like candy?}[$r$=-1] \\ 
\hline
i just got done watching a horror movie & i just got done watching a horror movie \\ 
\textcolor{blue}{i rather read, i have read about 20 books this year.}[$r$=+1] & \textcolor{red}{do you have any hobbies?}[$r$=-1] \\ 
\hline
wow! i do love a good horror movie. loving this cooler weather & wow! i do love a good horror movie. loving this cooler weather \\ 
\textcolor{blue}{but a good movie is always good.}[$r$=+1] & \textcolor{red}{good job! if you live to 100 like me, you will need all that learning.}[$r$=-1]\\ 
\hline
yes! my son is in junior high and i just started letting him watch them & yes! my son is in junior high and i just started letting him watch them \\ 
\textcolor{blue}{i work in the movies as well.}[$r$=+1] & \textcolor{red}{what a nice gesture. i take my dog to compete in agility classes.}[$r$=-1] \\ 
\hline
neat!! i used to work in the human services field & neat!! i used to work in the human services field \\ 
\textcolor{blue}{yes it is neat, i stunt double, it is so much fun and hard work.}[$r$=+1] & \textcolor{red}{you work very hard. i would like to do a handstand. can you teach it?}[$r$=-1] \\
\hline
yes i bet you can get hurt. my wife works and i stay at home & yes i bet you can get hurt. my wife works and i stay at home \\ 
\textcolor{blue}{nice, i only have one parent so now i help out my mom.}[$r$=+1] & \textcolor{red}{yes i do, red is one of my favorite colors}[$r$=-1] \\ 
\hline
i bet she appreciates that very much. & i bet she appreciates that very much. \\ 
\textcolor{blue}{she raised me right, i'm just like her.}[$r$=+1] & \textcolor{red}{haha, it is definitely attention grabbing!}[$r$=-1] \\ 
\hline
my dad was always busy working at home depot & my dad was always busy working at home depot \\ 
\textcolor{blue}{now that i am older home depot is my toy r us.}[$r$=+1] & \textcolor{red}{i bet there will be time to figure it out. what are your interests?}[$r$=-1] \\ 
\hline
\end{tabular}
\caption{Modified dialogue from the Persona-Chat dataset \cite{ZhangEtAl2018} with our proposed rewards: $r$=+1 means a human-like sentence and $r$=-1 means non-human like. The latter sentences, in red, are sampled randomly from different dialogues in the same dataset}
\label{rewards}
\end{center}
\end{table*}

\subsection{Human-Likeness Rewards}
\label{humanrewards}
Reward functions in reinforcement learning dialogue agents is often a difficult aspect. We propose to derive the rewards from human-human dialogues by assigning positive values to contextualised responses seen in the data, and negative values to randomly generated responses due to lacking coherence (also referred to as `non-human-like responses') -- see example in Table~\ref{rewards}. Thus, an episode or dialogue reward can be computed as $R_i=\sum_{j=1}^N r_j^i(a)$, where $i$ is the dialogue in focus, $j$ the dialogue turn in focus, and $r_j^i(a)$ is given according to
\begin{equation}\nonumber
  r^i_j(a)\mbox{=}\begin{cases}
    +1, & \text{if $a$ is a human response in dialogue-turn $i,j$}.\\
    -1, & \text{if $a$ is human but randomly chosen (incoherent)}.
  \end{cases}
\end{equation}


\subsection{Policy Learning}
Our Deep Reinforcement Learning (DRL) agents aim to maximise their cumulative reward overtime according to 
\begin{equation}\nonumber
Q^*(s,a;\theta)=\max_{\pi_\theta} {\mathbb E}[r_t+\gamma r_{t+1}+\gamma^2 r_{t+2}+\cdots|s,a,\pi_\theta], 
\end{equation}
where $r$ is the numerical reward given at time step $t$ for choosing action $a$ in state $s$, $\gamma$ is a discounting factor, and $Q^*(s,a;\theta)$ is the optimal action-value function using weights $\theta$ in a neural network. During training, a DRL agent will choose actions in a probabilistic manner in order to explore new $(s,a)$ pairs for discovering better rewards or to exploit already learnt values---with a reduced level of exploration overtime and an increased level of exploitation over time. During testing, a DRL agent will choose the best actions $a^*$ according to 
\begin{equation}\nonumber
\pi^*_\theta(s)=\arg\max_{a \in A} Q^*(s,a;\theta).
\end{equation}

Our DRL agents implement the procedure above using a generalisation of the DQN method \cite{MnihKSRVBGRFOPB15}---see Algorithm~\ref{ChatDQN}. After initialising replay memory $D$, dialogue history $H$, action-value function $Q$ and target action-value function $\hat{Q}$, we sample a training dialogue from our data of human-human conversations (lines 1-4). A human starts the conversation, which is mapped to its corresponding sentence embedding representation (lines 5-6). Then a set of candidate responses is generated including (1) the true human response and (2) randomly chosen responses (distractors). The candidate responses are clustered as described in Section~
\ref{clusteredactions} and the resulting actions are taken into account by the agent for action selection (lines 8-10). Once an action is chosen, it is conveyed to the environment, a reward is observed as described in Section~\ref{humanrewards}, and the agent's partner response is observed as well in order to update the dialogue history $H$ (lines 11-14). With such an update, the new sentence  embedding representation is generated from $H$ in order to update the replay memory $D$ with learning experience $(s,a,r,s')$ (lines 15-16). Then a minibatch of experiences $MB=(s_j,a_j,r_j,s_j')$ is sampled from $D$ in order to update the weights $\theta$ according to the error derived from the difference between the target value $y_j$ and the predicted value $Q(s,a;\theta)$ (see lines 18 and 20), which is based on the following loss function: 
\begin{equation}\nonumber
L(\theta_j)={\mathbb E}_{MB} \left( r+\gamma \max_{a'} \hat{Q}(s',a';\hat{\theta_j})-Q(s,a;\theta_j) \right)^2.
\end{equation}
The target action-value function $\hat{Q}$ and state $s$ are updated accordingly (lines 21-22), and this iterative procedure continues until convergence.

\begin{algorithm} [t!]
\caption{\label{ChatDQN} ChatDQN Learning}\label{ndqn} 
\begin{algorithmic}[1]
\State Initialise Deep Q-Networks with replay memory $D$, dialogue history $H$,  action-value function $Q$ with random weights $\theta$, and target action-value functions $\hat{Q}$ with $\hat{\theta}=\theta$
\State Initialise clustering model from training dialogue data
\Repeat
   \State Sample a training dialogue (human-human sentences)
   \State Append first sentence to dialogue history $H$
   \State $s=$ sentence embedding representation of $H$
   \Repeat 
      \State Generate {\it noisy} candidate response sentences
      \State $A=\text{cluster IDs of candidate response sentences}$
      \State $a= 
\begin{cases}
    rand_{a \in A} \text{ if } \mbox{random number} \le \epsilon\\
    \max_{a \in A} Q(s,a;\theta)  \text{  otherwise}
\end{cases}$
      \State Execute chosen clustered action $a$
      \State Observe human-likeness dialogue reward $r$
      \State Observe environment response (agent's partner)
      \State Append agent and environment responses to $H$
      \State $s'=$ sentence embedding representation of $H$
      \State Append transition ($s,a,r,s'$) to $D$
      \State Sample random minibatch $(s_j,a_j,r_j,s'_j)$ from $D$
      \State $y_j= 
\begin{cases}
    r_j \text{ if } \mbox{final step of episode}\\
    r_j + \gamma \max_{a' \in A} \hat{Q}(s',a';\hat{\theta})              & \text{otherwise}
\end{cases}$
      \State Set $err=\left( y_j-Q(s,a;\theta) \right)^2$ 
      \State Gradient descent step on $err$ with respect to $\theta$
      \State Reset $\hat{Q}=Q$ every $C$ steps
      \State $s \leftarrow$ $s'$
   \Until {end of dialogue}
   \State Reset dialogue history $H$
\Until convergence
\end{algorithmic}
\end{algorithm}

\section{Experiments and Results}
\label{experiments}

\subsection{Data}
We used data from the {\it Persona-Chat} data set\footnote{Data set downloaded from \url{http://parl.ai/} on 18 May 2018 \cite{MillerFBBFLPW17}}, which includes 17,877 dialogues for training (131,431 turns) and 999 dialogues for testing (7,793 turns). They represent averages of 7.35 and 7.8 dialogue turns for training and testing, respectively---see example dialogue in Table~\ref{rewards}. The vocabulary size in the entire data set contains 19,667 unique words. 



\subsection{Experimental Setting}
\label{expsetting}
\begin{figure*}[t]
\centering
\subfigure[100 clusters of training sentences]{\includegraphics[width=130mm]{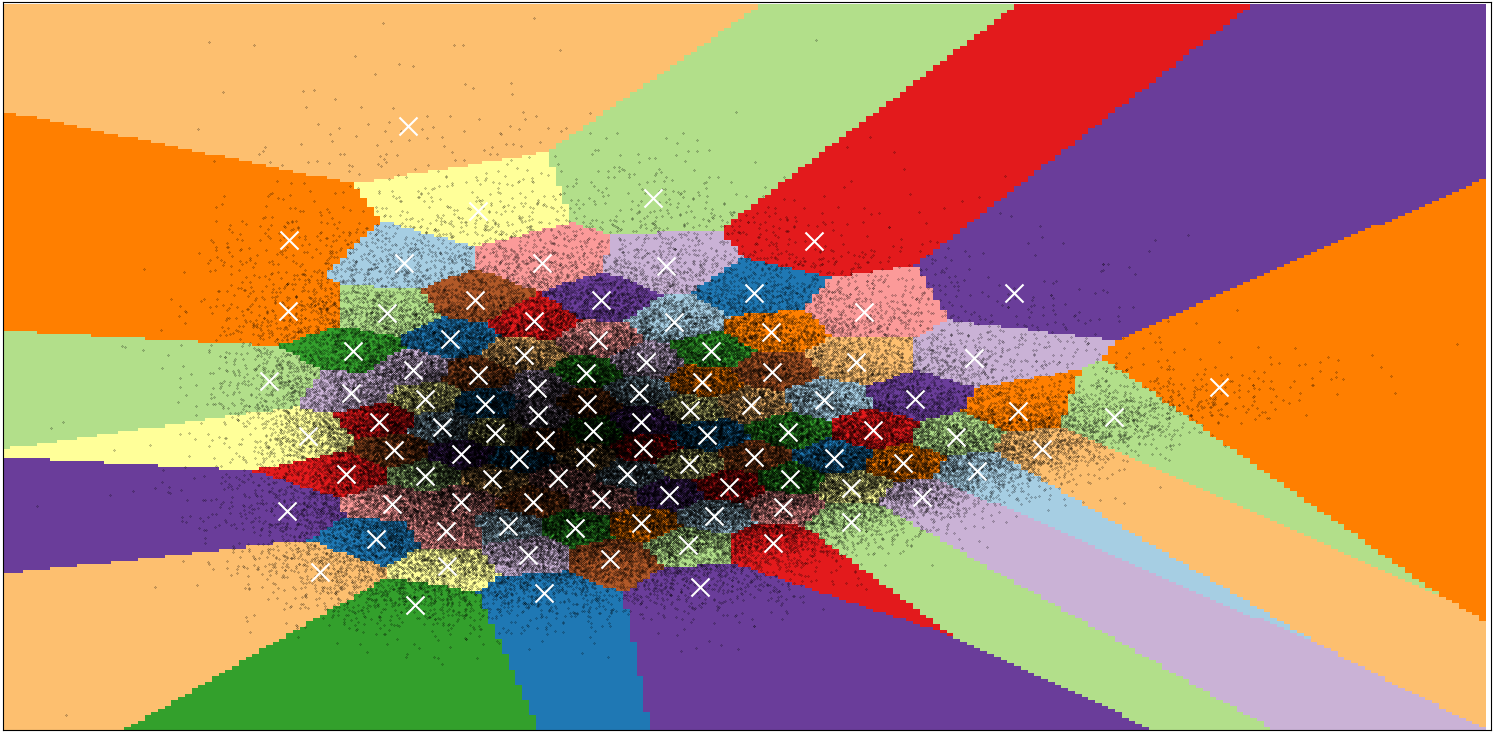}}
\subfigure[20 clusters of training dialogues]{\includegraphics[width=130mm]{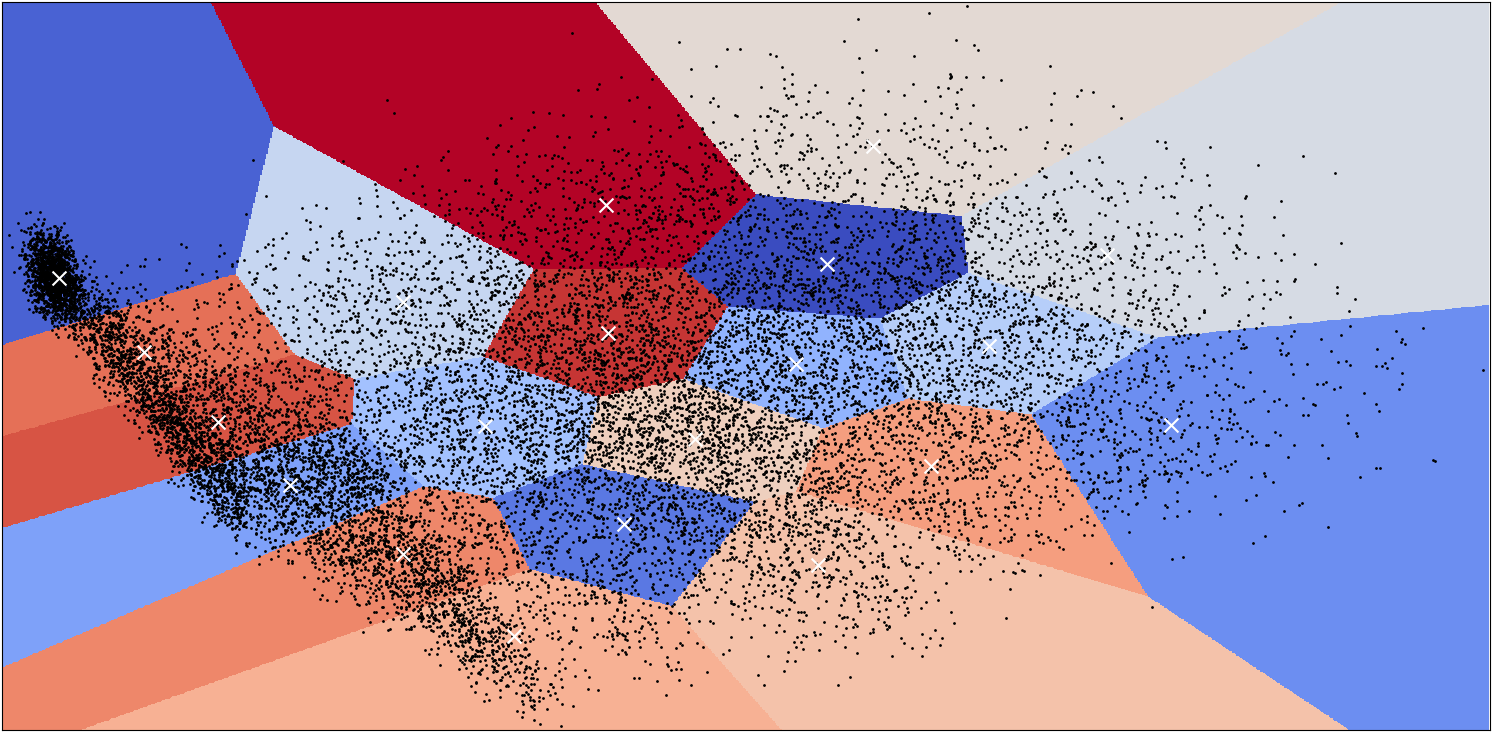}}
\caption{Example clusters of our training data using Principal Component Analysis \cite{PPCA} for visualisations in 2D -- black dots represent sentences or dialogues}
\label{exampleclusters}
\end{figure*}

To analyse the performance of our ChatDQN agents we use subsets of training data vs. the entire training data set. The former are automatically generated by using sentence vectors to represent the features of each dialogue---as described in Section~\ref{clusteredactions}. Similarly, 
the agents' states are modelled using sentence vectors of the dialogue history with the pretrained coefficients of the Glove model \cite{PenningtonSM14}. In all our experiments we use the following neural network architecture\footnote{Other hyperparameters include embedding batch size=128, dropout=0.2, latent dimensionality=256, discount factor=0.99, size of candidate responses=3, max. number of sentence vectors in $H$=50, burning steps=3K, memory size=10K, target model update (C)=10K, learning steps=50K, test steps=100K. The number of parameters in our neural nets with 100 and 300 sentence vector dimensions corresponds to 4.4 and 12.1 million, respectively.}: 
\begin{itemize}
\item mean word vectors, one per sentence, in the input layer (maximum number of vectors=50, with zero-padding) -- each word vector of 100 or 300 embedding size, 
\item two Gated Recurrent Unit (GRU) \cite{choEtAlEMNLP2014} layers with latent dimensionality of 256, and 
\item fully connected layer with number of nodes=the number of clusters, i.e. each cluster corresponding to one action. 
\end{itemize}
While a small number of sentence clusters could result in actions being assigned to potentially the same cluster, a larger number of sentence clusters would mitigate the problem, but the larger the number of clusters the larger the computational expense---i.e. more parameters in the neural network. Figure~\ref{exampleclusters}(a) shows an example of our sentence clustering using 100 clusters on our training data. A manual inspection showed that greeting sentences were mostly assigned to the same cluster, and questions expressing preferences (e.g. What is your favourite X?) were also assigned to the same cluster. In this work we thus use a sentence clustering model with $k$=100 derived from our training data and prior to reinforcement learning\footnote{Each experiment in this paper was ran on a GPU Tesla K80 using the following libraries: Keras (\url{https://github.com/keras-team/keras}), OpenAI (\url{https://github.com/openai}) and Keras-RL (\url{https://github.com/keras-rl/keras-rl}).}. In addition, we trained a second clustering model to analyse our experiments using different data splits, where instead of clustering sentences we cluster dialogues. Given that we represent a sentence using a mean word vector, a dialogue can thus be represented by a group of sentence vectors. Figure~\ref{exampleclusters}(b) shows an example of our dialogue clustering using 20 clusters on our training data.

Notice that while previous related works in task-oriented DRL-based agents typically use a user simulator, this paper does not use a simulator. Instead, we use the dataset of human-human dialogues directly and substitute one partner conversant in the dialogues by a DRL agent. The goal of the agent is to choose the human-generated sentences (actions) out of a set of candidate responses.

\begin{figure*}[t]
\begin{flushleft}
\subfigure[ChatDQN agents using data splits 0 to 4 (from left to right)]{
\includegraphics[width=35mm]{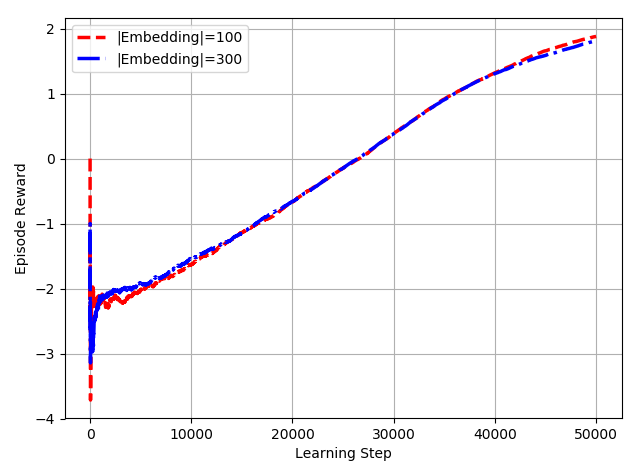}
\includegraphics[width=35mm]{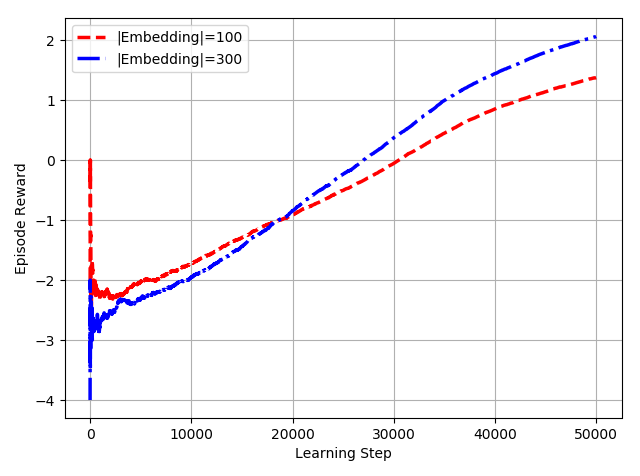}
\includegraphics[width=35mm]{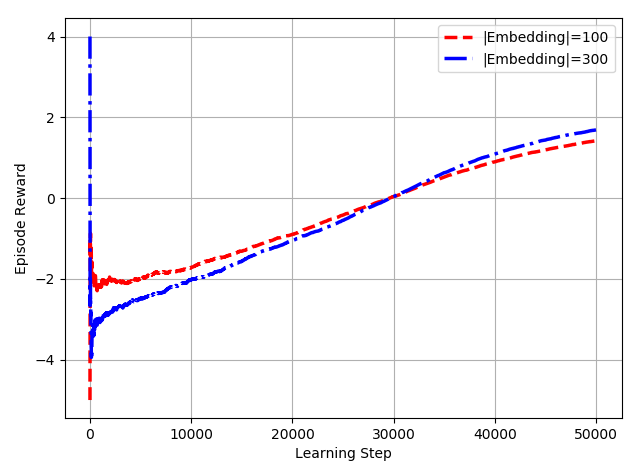}
\includegraphics[width=35mm]{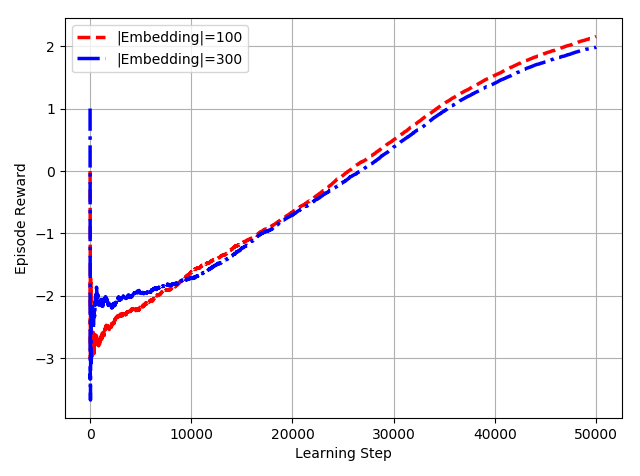}
\includegraphics[width=35mm]{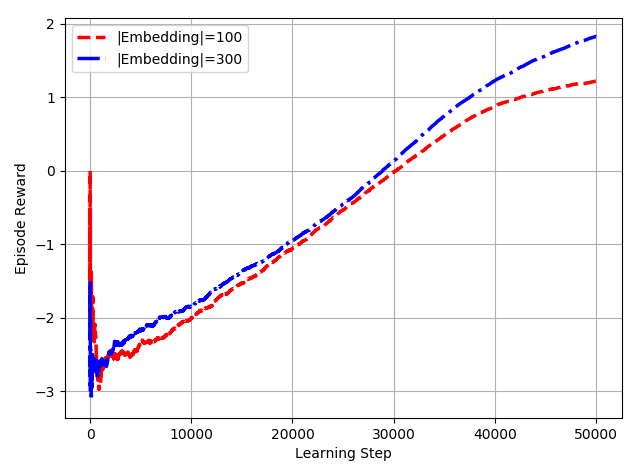}
}
\subfigure[ChatDQN agents using data splits 5 to 9 (from left to right)]{
\includegraphics[width=35mm]{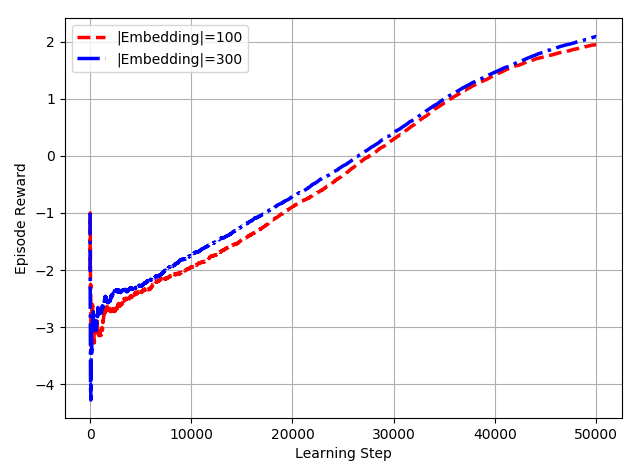}
\includegraphics[width=35mm]{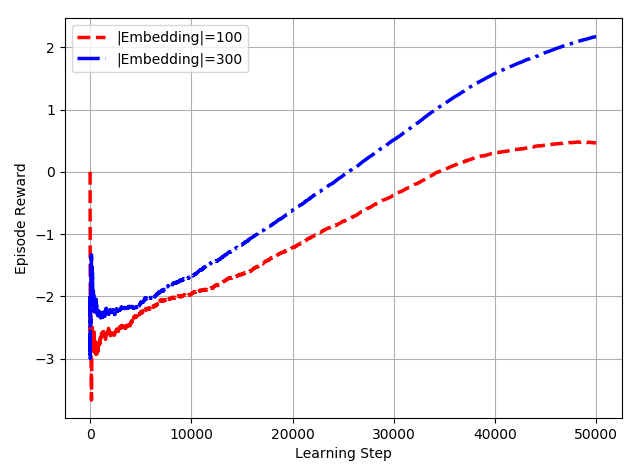}
\includegraphics[width=35mm]{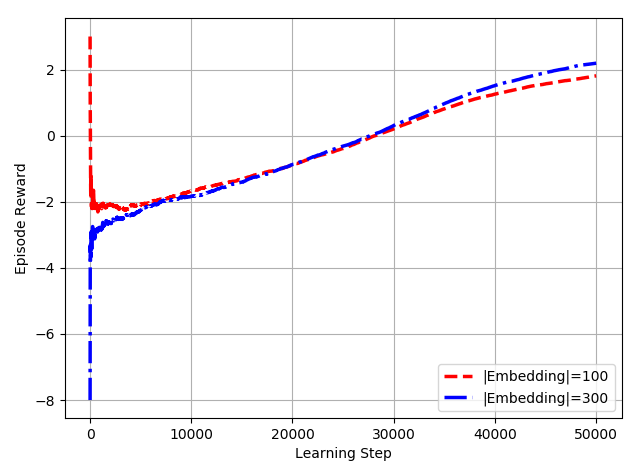}
\includegraphics[width=35mm]{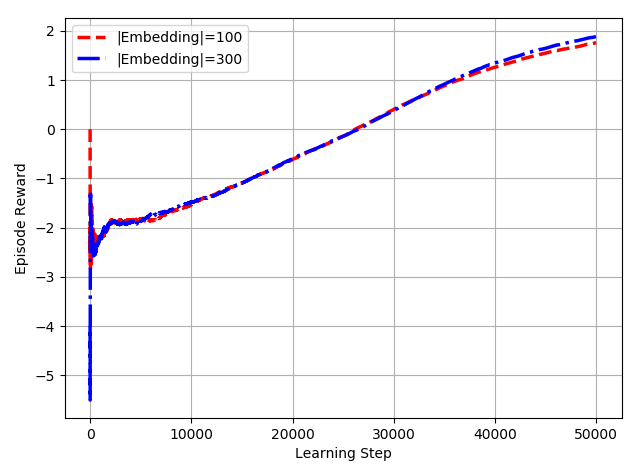}
\includegraphics[width=35mm]{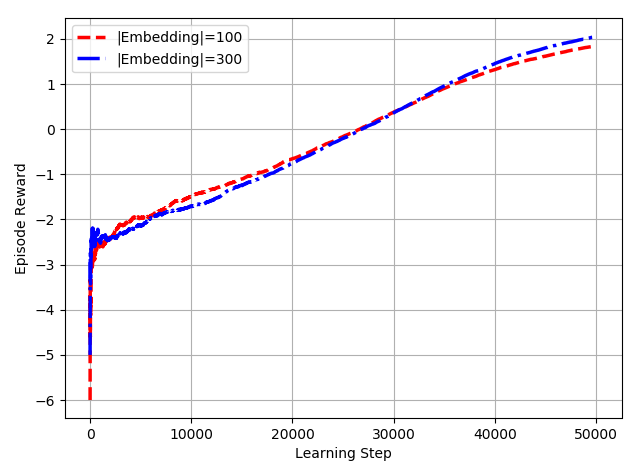}
}
\subfigure[ChatDQN agents using data splits 10 to 14 (from left to right)]{
\includegraphics[width=35mm]{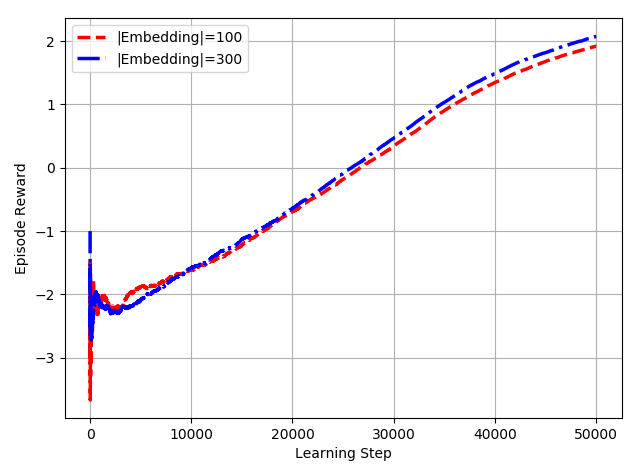}
\includegraphics[width=35mm]{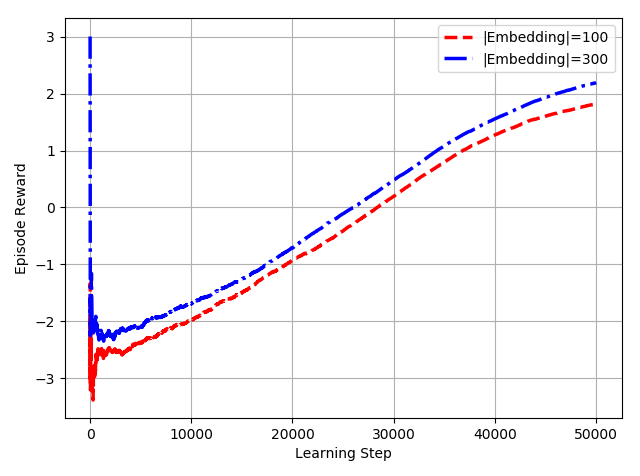}
\includegraphics[width=35mm]{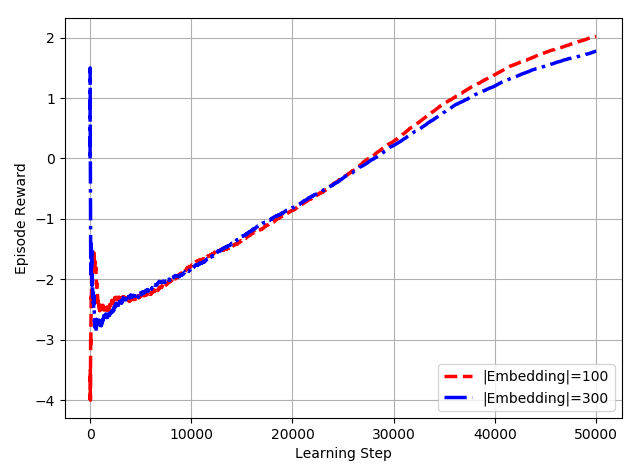}
\includegraphics[width=35mm]{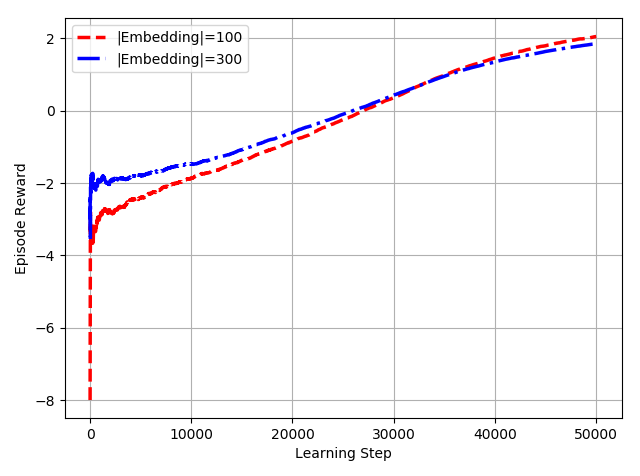}
\includegraphics[width=35mm]{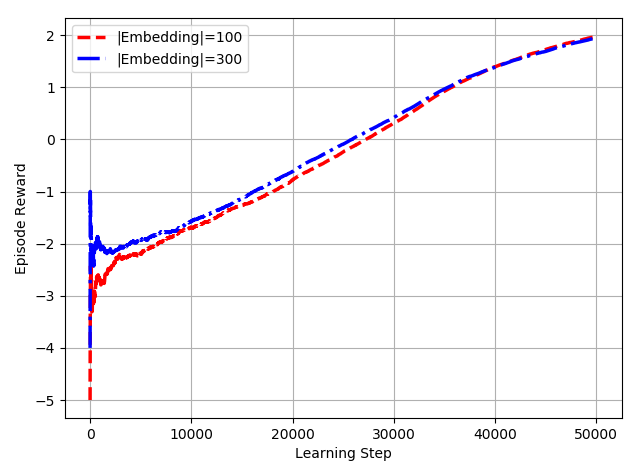}
}
\subfigure[ChatDQN agents using data splits 15 to 19 (from left to right)]{
\includegraphics[width=35mm]{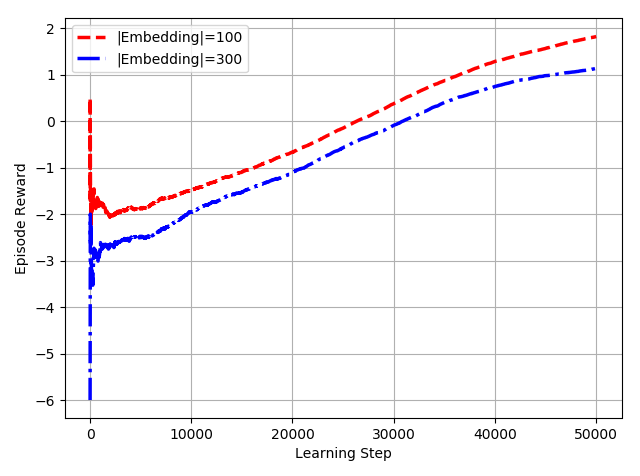}
\includegraphics[width=35mm]{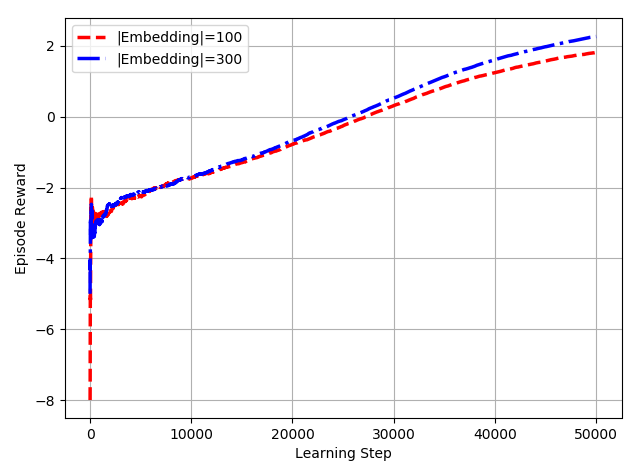}
\includegraphics[width=35mm]{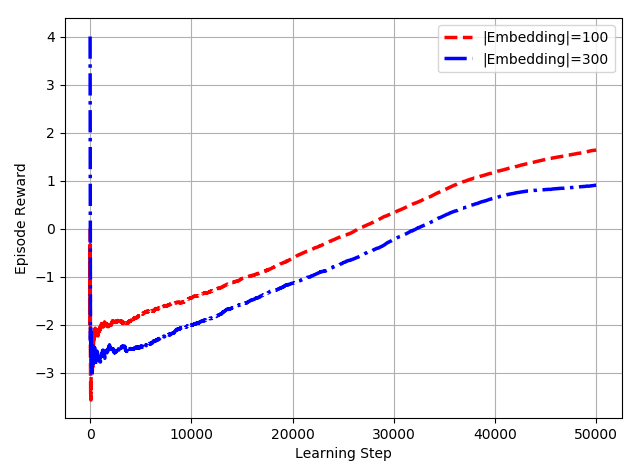}
\includegraphics[width=35mm]{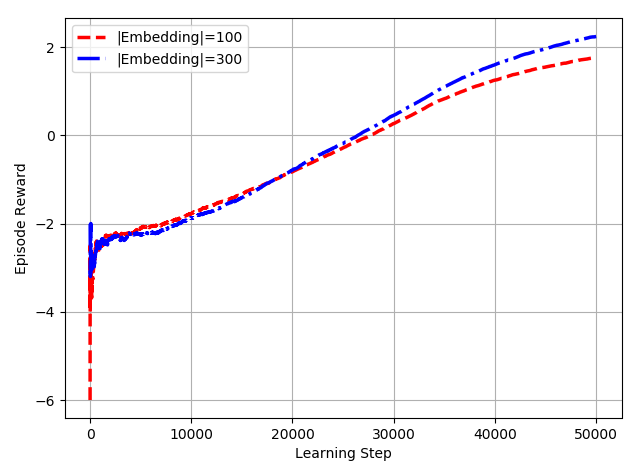}
\includegraphics[width=35mm]{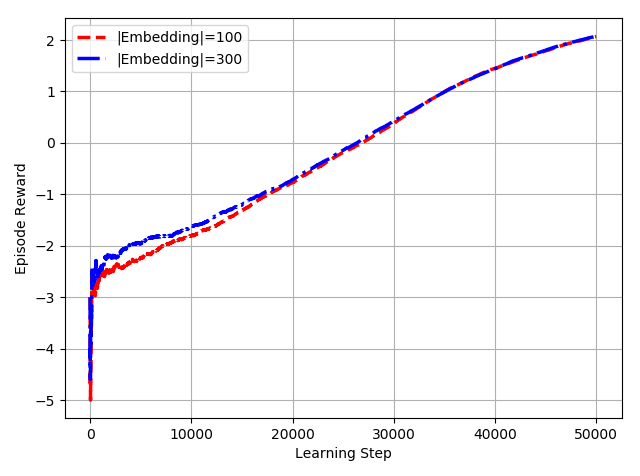}
}
\vspace{-0.5cm}
\end{flushleft}
\caption{Training performance of ChatDQN agents using different data splits of dialogues---see text for details}
\label{learningcurves}
\end{figure*}

\begin{table*}[t!]
      \centering
      \caption{Average reward results of ChatDQN agents 0 to 20 trained with different data splits and size of sentence embedding (42 agents in total), where $\dagger$ denotes significant difference (at $p=0.05$) using a two-tailed Wilcoxon Signed Rank Test}
	  \label{testresults100}
\begin{tabular}{|c|c|c|c||c|c|c|c|} 
\hline
\multicolumn{4}{|c}{$\mid$Embedding$\mid$=100} & \multicolumn{4}{c|}{$\mid$Embedding$\mid$=300} \\
\cline{1-8}
Data Split & Training & Testing on the & Testing on the & Data Split & Training & Testing on the & Testing on the\\
($\vert$dialogues$\vert$) &          & Training Set & Test Set & ($\vert$dialogues$\vert$) & & Training Set & Test Set\\
\hline
\hline
0 (861) & 1.8778 & 3.7711 & -1.1708 & 0 (1000) & 1.8168 & 3.6785 & -0.8618 \\
1 (902) & 1.3751 & 3.1663 & -1.7006 & 1 (850) & 2.0622 & 4.4598 & -1.8688 \\
2 (907) & 1.4194 & 3.1579 & -0.9723 & 2 (1010) & 1.6896 & 3.6724 & -1.4282 \\
3 (785) & 2.1532 & 4.2508 & -1.3444 & 3 (1029) & 1.9845 & 4.0136 & -0.6109 \\
4 (1046) & 1.2204 & 2.1581 & -1.5633 & 4 (951) & 1.8255 & 4.0423 & -1.4448 \\
5 (767) & 1.9456 & 3.9017 & -1.2123 & 5 (832) & 2.0860 & 4.2182 & -0.8277 \\
6 (1053) & 0.4621 & 0.1370 & -1.8443 & 6 (815) & 2.1735 & 4.2592 & -1.5193 \\
7 (968) & 1.8090 & 3.8368 & -1.1137 & 7 (891) & 2.1921 & 4.5799 & -1.4233 \\
8 (858) & 1.7608 & 3.5531 & -1.6678 & 8 (905) & 1.8835 & 3.8337 & -0.6628 \\
9 (826) & 1.8431 & 3.6254 & -1.0919 & 9 (892) & 2.0521 & 4.1882 & -1.5267 \\
10 (818) & 1.9188 & 3.8629 & -0.5394 & 10 (835) & 2.0709 & 4.2852 & -0.8831\\
11 (944) & 1.8212 & 3.5724 & -1.7020 & 11 (873) & 2.1902 & 4.4848 & -1.3329\\
12 (873) & 2.0195 & 4.1895 & -1.3456 & 12 (948) & 1.7761 & 3.7927 & -1.6167\\
13 (895) & 2.0515 & 4.1873 & -1.8034 & 13 (932) & 1.8563 & 3.6208 & -1.5149\\
14 (863) & 1.9722 & 4.1479 & -1.3244 & 14 (812) & 1.9486 & 4.0347 & -1.5866\\
15 (842) & 1.8214 & 3.8942 & -0.8921 & 15 (880) & 1.1338 & 2.4880 & -1.4084\\
16 (837) & 1.8162 & 3.8817 & -1.3784 & 16 (787) & 2.2628 & 4.5583 & -1.4290\\
17 (958) & 1.6373 & 3.3373 & -0.7726 & 17 (994) & 0.9038 & 1.5106 & -1.5925\\
18 (1012) & 1.7631 & 3.6279 & -1.2690 & 18 (853) & 2.2405 & 4.4716 & -1.4231\\
19 (862) & 2.0683 & 4.2026 & -1.5901 & 19 (788) & 2.0686 & 4.2219 & -0.9594\\
20 (17877) & -0.4138 & -1.2473 & -1.9684 & 20 (17877) & -0.3516 & -0.3490 & -2.0870 \\
Average$^{0-20}$ & 1.6353 & 3.2959$\dagger$ & -1.3461 & Average$^{0-20}$ & {\bf 1.8031} & {\bf 3.7174}$\dagger$ & {\bf -1.3337}\\
Sum$^{0-20}$ & 34.3419 & 69.2146 & -28.2674 & Sum$^{0-20}$ & {\bf 37.8656} & {\bf 78.0653} & {\bf -28.0079}\\
\hline
Upper Bound & 7.1810 & 7.1810 & 7.5942 & Upper Bound & 7.1810 & 7.1810 & 7.5942 \\ 
Lower Bound & -7.2834 & -7.2834 & -7.7276 & Lower Bound & -7.2834 & -7.2834 & -7.7276 \\ 
Random Sel. & -2.4139 & -2.4139 & -2.5526 & Random Sel. & -2.4139 & -2.4139 & -2.5526 \\ 
\hline
\end{tabular}
\end{table*}


\subsection{Experimental Results}
The plots in Figure~\ref{learningcurves} show the training performance of our ChatDQN agents---all using 100 clustered actions. Each plot contains two learning curves, one per agent, where each agent uses a different sentence embedding size (100 or 300 dimensions). In addition, each plot uses an automatically generated data split according to our clustered dialogues. These plots show evidence that all agents indeed improve their behaviour over time even when they use only 100 actions. This can be observed from their average episode rewards, the higher the better in all learning curves. 
From a visual inspection, we can observe that the agents using either embedding size (100 or 300) perform rather equivalently but with a small trend for 300 dimensions to dominate its counterpart -- more on this below.

\begin{figure}[t!]
\centering
{\includegraphics[width=55mm]{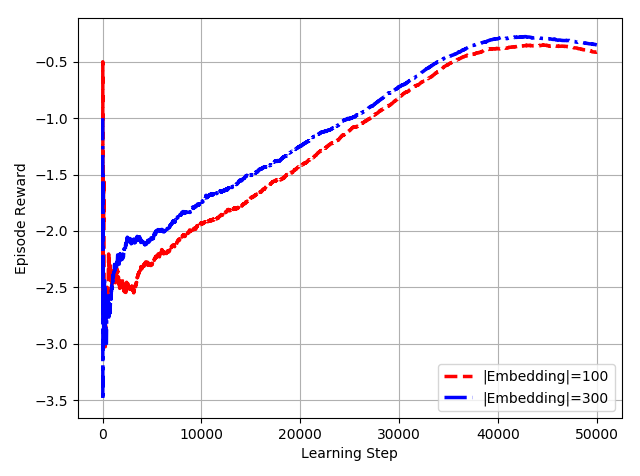}}
\caption{Training performance of our ChatDQN agents using all training dialogues and two sentence embedding sizes}
\label{learningcurves2}
\end{figure}

The performance of our ChatDQN agents using all training dialogues is shown in Figure~\ref{learningcurves2}. It can be noted that in contrast to the previous agents where their improvement in average reward reached values of around 2, the performance in these agents was lower (with average episode reward $<0$). We attribute this to the larger amount of variation exhibited from about 1K dialogues to 17.8K dialogues.

We analysed the performance of our agents further by using a test set of totally unseen dialogues during training. Table~\ref{testresults100} summarises our results, where we can note that the larger sentence embedding size (300) generally performed better. While a significant difference (according to a two-tailed Wilcoxon Singed Rank Test) at $p=0.05$ was identified in testing on the training set, {\it no significant difference} was found in performance during testing on the test set. These results could be confirmed in other datasets and/or settings in future work. In addition, we can observe that the ChatDQN agents trained using all data (agents with id=20) were not able to achieve as good performance than those agents using smaller data splits. Our results thus reveal that training chatbots on some sort of domains (groups of dialogues automatically discovered in our case), is useful for improved performance.

\section{Analysis of Human-Likeness Rewards}
\label{rewards}
We employ the algorithm of \cite{CuayahuitlEtAl2018neurips} for extending a dataset of human-human dialogues with distorted dialogues. The latter include varying amounts of distortions, i.e. different degrees of human-likeness. We use such data for training and testing reward prediction models in order to analise the goodness of our proposed reward function. Given extended dataset $\mathcal{\hat{D}}=\{(\hat{d}_1,y_1),\dots,(\hat{d}_N,y_N)\}$ with (noisy) dialogue histories $\hat{d}_i$, the goal is to predict dialogue scores $y_i$ as accurately as possible. We represent a dialogue history via its sentence vectors as in  Deep Averaging Networks \cite{IyyerMBD15}, where sentences are represented with numerical feature vectors denoted as ${\bf x}=\{x_1,...,x_{|{\bf x}|}\}$. In this way, a set of word sequences $s^i_j$ in dialogue-sentence pair $i,j$ is mapped to feature vectors 
\begin{equation}\nonumber
{\bf x}^i_j=\frac{1}{N^i_j}\sum_{k=1}^{N^i_j} c^i_{j,k},
\end{equation} 
where $c^i_{j,k}$ is the vector of coefficients of word $k$, part of sentence $j$ in dialogue $i$, and $N^i_j$ is the number of words in the sentence in focus.

Assuming that vector ${\bf Y}=\{y_1,...,y_{|{\bf Y}|}\}$ is the set of target labels---generated as described in the dialogue generation algorithm of \cite{CuayahuitlEtAl2018neurips}, and using the same test data as the previous section. In this way, dataset $\mathcal{D}^{train}=({\bf X}^{train},{\bf Y}^{train})$ is used for training neural regression models using varying amounts of dialogue history, and dataset $\mathcal{D}^{test}=({\bf X}^{test},{\bf Y}^{test})$ is used for testing the learnt models.

Our experiments use a 2-layer Gated Recurrent Unit (GRU) neural network \cite{choEtAlEMNLP2014}, similar to the one in Section¬\ref{expsetting} but including Batch Normalisation \cite{IoffeS15} between hidden layers. 

\begin{figure}[t]
\centering
\includegraphics[width=80mm]{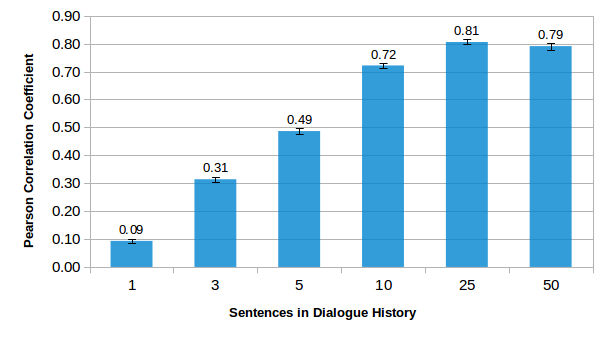}
\caption{Bar plot showing the performance of our dialogue reward predictors using different amounts of dialogue history (from 1 sentence to 50 sentences). Each bar reports an average Pearson correlation score over 10 runs, where the coefficients report the correlation between true dialogue rewards and predicted dialogue rewards in our test data}
\label{barplot}
\end{figure}

We trained neural networks for six different lengths of dialogue history, ranging from 1 sentence to 50 sentences. Each length size involved a separate neural network, trained 10 times in order to report results over multiple runs. Figure~\ref{barplot} reports the average Pearson correlation coefficient---between true dialogue rewards and predicted dialogue rewards---for each length size. It can be observed that short dialogue histories contribute to obtain weak correlations, and that longer dialogue histories ($\geq 10$ sentences) contribute to obtain strong correlations. It can also be observed that the longest history may not be the best choice of length size, the network using $25$ sentences achieved the best results. From these results we can conclude that our proposed human-likeness rewards---with lengthy dialogue histories---can be used for training future neural-based chatbots.


\section{Conclusion and Future Work}
\label{conclusion}
This paper presents a novel approach for training Deep Reinforcement Learning (DRL) chatbots, which uses clustered actions and rewards derived from human-human dialogues without any manual annotations. The task of the agents is to learn to choose human-like actions (sentences) out of candidate responses including human generated and randomly generated sentences. In our proposed rewards we assume that the latter are generally incoherent throughout the dialogue history. Experimental results using chitchat data report that DRL agents learn reasonable policies using training dialogues, but their generalisation ability in a test set of unseen dialogues remains a key challenge for future research in this field. In addition, we found the following: (a) that sentence embedding sizes of 100 and 300 perform equivalently on test data; (b) that training agents using larger amounts of training can deteriorate performance than training with smaller amounts; and (c) that our proposed dialogue rewards can be predicted with strong correlation (between true and predicted rewards) by using neural-based regressors with lengthy dialogue histories of $\ge$ 10 sentences (25 sentences was the best in our experiments).

Future work can explore the following avenues. First, confirm these findings with other datasets and settings in order to draw even stronger conclusions. Second, investigate further the proposed approach for improved generalisation in test data. For example, other methods of feature extraction, clustering algorithms, distance metrics, policy learning algorithms, architectures, and a comparison of reward functions can be explored. Last but not least, combine the proposed learning approach with more knowledge intensive resources \cite{ManningEtAl2014acl,Dethlefs17} such as semantic parsers, coreference resolution, among others.

\bibliographystyle{IEEEbib}
\bibliography{refs}

\begin{thebibliography}{10}

\bibitem{GroszS86}
Barbara~J. Grosz and Candace~L. Sidner,
\newblock ``Attention, intentions, and the structure of discourse,''
\newblock {\em Computational Linguistics}, vol. 12, no. 3, pp. 175--204, 1986.

\bibitem{SuttonB2018}
Richard~S. Sutton and Andrew~G. Barto,
\newblock {\em Reinforcement learning - an introduction},
\newblock Adaptive computation and machine learning. {MIT} Press, 2nd edition
  edition, 2018.

\bibitem{HastieTF09}
Trevor Hastie, Robert Tibshirani, and Jerome~H. Friedman,
\newblock {\em The elements of statistical learning: data mining, inference,
  and prediction, 2nd Edition},
\newblock Springer series in statistics. Springer, 2009.

\bibitem{LeCunBH15}
Yann LeCun, Yoshua Bengio, and Geoffrey~E. Hinton,
\newblock ``Deep learning,''
\newblock {\em Nature}, vol. 521, no. 7553, pp. 436--444, 2015.

\bibitem{CasanuevaBSURTG18}
I{\~{n}}igo Casanueva, Pawel Budzianowski, Pei{-}Hao Su, Stefan Ultes,
  Lina~Maria Rojas{-}Barahona, Bo{-}Hsiang Tseng, and Milica Gasic,
\newblock ``Feudal reinforcement learning for dialogue management in large
  domains,''
\newblock in {\em {NAACL-HLT}}, 2018.

\bibitem{Cuayahuitl16}
Heriberto Cuay{\'{a}}huitl,
\newblock ``Simple{DS}: {A} simple deep reinforcement learning dialogue
  system,''
\newblock {\em CoRR}, vol. abs/1601.04574, 2016.

\bibitem{CuayahuitlYWC17}
Heriberto Cuay{\'{a}}huitl, Seunghak Yu, Ashley Williamson, and Jacob Carse,
\newblock ``Scaling up deep reinforcement learning for multi-domain dialogue
  systems,''
\newblock in {\em {IJCNN}}, 2017.

\bibitem{CuayahuitlY17}
Heriberto Cuay{\'{a}}huitl and Seunghak Yu,
\newblock ``Deep reinforcement learning of dialogue policies with less weight
  updates,''
\newblock in {\em {INTERSPEECH}}, 2017.

\bibitem{WilliamsAZ17}
Jason~D. Williams, Kavosh Asadi, and Geoffrey Zweig,
\newblock ``Hybrid code networks: practical and efficient end-to-end dialog
  control with supervised and reinforcement learning,''
\newblock in {\em {ACL}}, 2017.

\bibitem{PengLLGCLW17}
Baolin Peng, Xiujun Li, Lihong Li, Jianfeng Gao, Asli {\c{C}}elikyilmaz,
  Sungjin Lee, and Kam{-}Fai Wong,
\newblock ``Composite task-completion dialogue policy learning via hierarchical
  deep reinforcement learning,''
\newblock in {\em {EMNLP}}, 2017.

\bibitem{LiMSJRJ16}
Jiwei Li, Will Monroe, Alan Ritter, Dan Jurafsky, Michel Galley, and Jianfeng
  Gao,
\newblock ``Deep reinforcement learning for dialogue generation,''
\newblock in {\em {EMNLP}}, 2016.

\bibitem{LiMSJRJ17}
Jiwei Li, Will Monroe, Tianlin Shi, S{\'{e}}bastien Jean, Alan Ritter, and Dan
  Jurafsky,
\newblock ``Adversarial learning for neural dialogue generation,''
\newblock in {\em {EMNLP}}, 2017.

\bibitem{SerbanEtAl2018}
Iulian~Vlad Serban, Chinnadhurai Sankar, Mathieu Germain, Saizheng Zhang,
  Zhouhan Lin, Sandeep Subramanian, Taesup Kim, Michael Pieper, Sarath Chandar,
  Nan~Rosemary Ke, Sai Rajeswar, Alexandre de~Br{\'{e}}bisson, Jose M.~R.
  Sotelo, Dendi Suhubdy, Vincent Michalski, Alexandre Nguyen, Joelle Pineau,
  and Yoshua Bengio,
\newblock ``A deep reinforcement learning chatbot (short version),''
\newblock {\em CoRR}, vol. abs/1801.06700, 2018.

\bibitem{SankarRavi2018neurips}
Chinnadhurai Sankar and Sujith Ravi,
\newblock ``Modeling non-goal oriented dialog with discrete attributes,''
\newblock in {\em NeurIPS Workshop on Conversational AI: ``Today's Practice and
  Tomorrow`s Potential''}, 2018.

\bibitem{abs-1804-02504}
Chih{-}Wei Lee, Yau{-}Shian Wang, Tsung{-}Yuan Hsu, Kuan{-}Yu Chen, Hung{-}yi
  Lee, and Lin{-}Shan Lee,
\newblock ``Scalable sentiment for sequence-to-sequence chatbot response with
  performance analysis,''
\newblock {\em CoRR}, vol. abs/1804.02504, 2018.

\bibitem{VinyalsL15}
Oriol Vinyals and Quoc~V. Le,
\newblock ``A neural conversational model,''
\newblock {\em CoRR}, vol. abs/1506.05869, 2015.

\bibitem{SordoniGABJMNGD15}
Alessandro Sordoni, Michel Galley, Michael Auli, Chris Brockett, Yangfeng Ji,
  Margaret Mitchell, Jian{-}Yun Nie, Jianfeng Gao, and Bill Dolan,
\newblock ``A neural network approach to context-sensitive generation of
  conversational responses,''
\newblock in {\em {HLT-NAACL}}, 2015.

\bibitem{SerbanKTTZBC17}
Iulian~Vlad Serban, Tim Klinger, Gerald Tesauro, Kartik Talamadupula, Bowen
  Zhou, Yoshua Bengio, and Aaron~C. Courville,
\newblock ``Multiresolution recurrent neural networks: An application to
  dialogue response generation,''
\newblock in {\em {AAAI}}, 2017.

\bibitem{LiGBSGD16}
Jiwei Li, Michel Galley, Chris Brockett, Georgios~P. Spithourakis, Jianfeng
  Gao, and William~B. Dolan,
\newblock ``A persona-based neural conversation model,''
\newblock in {\em {ACL}}, 2016.

\bibitem{Wang2018}
Wenjie Wang, Minlie Huang, Xin-Shun Xu, Fumin Shen, and Liqiang Nie,
\newblock ``Chat more: Deepening and widening the chatting topic via a deep
  model,''
\newblock in {\em {SIGIR}}. 2018, ACM.

\bibitem{ZhangEtAl2018}
Saizheng Zhang, Emily Dinan, Jack Urbanek, Arthur Szlam, Douwe Kiela, and Jason
  Weston,
\newblock ``Personalizing dialogue agents: {I} have a dog, do you have pets
  too?,''
\newblock {\em CoRR}, vol. abs/1801.07243, 2018.

\bibitem{Yan18}
Rui Yan,
\newblock ``"chitty-chitty-chat bot": Deep learning for conversational {AI},''
\newblock in {\em {IJCAI}}, 2018.

\bibitem{LiuLSNCP16}
Chia{-}Wei Liu, Ryan Lowe, Iulian Serban, Michael Noseworthy, Laurent Charlin,
  and Joelle Pineau,
\newblock ``How {NOT} to evaluate your dialogue system: An empirical study of
  unsupervised evaluation metrics for dialogue response generation,''
\newblock in {\em {EMNLP}}, 2016.

\bibitem{MikolovSCCD13}
Tomas Mikolov, Ilya Sutskever, Kai Chen, Gregory~S. Corrado, and Jeffrey Dean,
\newblock ``Distributed representations of words and phrases and their
  compositionality,''
\newblock in {\em {NIPS}}, 2013.

\bibitem{PenningtonSM14}
Jeffrey Pennington, Richard Socher, and Christopher~D. Manning,
\newblock ``Glove: Global vectors for word representation,''
\newblock in {\em {EMNLP}}, 2014.

\bibitem{IyyerMBD15}
Mohit Iyyer, Varun Manjunatha, Jordan~L. Boyd{-}Graber, and Hal~Daum{\'{e}}
  III,
\newblock ``Deep unordered composition rivals syntactic methods for text
  classification,''
\newblock in {\em {ACL} {(1)}}, 2015.

\bibitem{ArthurV07}
David Arthur and Sergei Vassilvitskii,
\newblock ``K-means++: The advantages of careful seeding,''
\newblock in {\em {SODA}}. 2007, {SIAM}.

\bibitem{MnihKSRVBGRFOPB15}
Volodymyr Mnih, Koray Kavukcuoglu, David Silver, Andrei~A. Rusu, Joel Veness,
  Marc~G. Bellemare, Alex Graves, Martin~A. Riedmiller, Andreas Fidjeland,
  Georg Ostrovski, Stig Petersen, Charles Beattie, Amir Sadik, Ioannis
  Antonoglou, Helen King, Dharshan Kumaran, Daan Wierstra, Shane Legg, and
  Demis Hassabis,
\newblock ``Human-level control through deep reinforcement learning,''
\newblock {\em Nature}, vol. 518, no. 7540, 2015.

\bibitem{MillerFBBFLPW17}
Alexander~H. Miller, Will Feng, Dhruv Batra, Antoine Bordes, Adam Fisch, Jiasen
  Lu, Devi Parikh, and Jason Weston,
\newblock ``Parlai: {A} dialog research software platform,''
\newblock in {\em {EMNLP} (System Demonstrations)}, 2017.

\bibitem{choEtAlEMNLP2014}
Kyunghyun Cho, Bart van Merrienboer, Caglar Gulcehre, Dzmitry Bahdanau, Fethi
  Bougares, Holger Schwenk, and Yoshua Bengio,
\newblock ``Learning phrase representations using {RNN} encoder--decoder for
  statistical machine translation,''
\newblock in {\em {EMNLP}}. 2014, Association for Computational Linguistics.

\bibitem{PPCA}
M.~E. Tipping and Christopher Bishop,
\newblock ``Probabilistic principal component analysis,''
\newblock {\em Journal of the Royal Statistical Society, Series B}, vol. 21/3,
  pp. 611–622, January 1999.

\bibitem{CuayahuitlEtAl2018neurips}
Heriberto Cuay\'ahuitl, Seonghan Ryu, Donghyeon Lee, and Jihie Kim,
\newblock ``A study on dialogue reward prediction for open-ended conversational
  agents,''
\newblock in {\em NeurIPS Workshop on Conversational AI: ``Today's Practice and
  Tomorrow`s Potential''}, 2018.

\bibitem{IoffeS15}
Sergey Ioffe and Christian Szegedy,
\newblock ``Batch normalization: Accelerating deep network training by reducing
  internal covariate shift,''
\newblock in {\em International Conference on Machine Learning (ICML)}, 2015.

\bibitem{ManningEtAl2014acl}
Christopher~D. Manning, Mihai Surdeanu, John Bauer, Jenny Finkel, Steven~J.
  Bethard, and David McClosky,
\newblock ``The {Stanford} {CoreNLP} natural language processing toolkit,''
\newblock in {\em Association for Computational Linguistics (ACL) System
  Demonstrations}, 2014.

\bibitem{Dethlefs17}
Nina Dethlefs,
\newblock ``Domain transfer for deep natural language generation from abstract
  meaning representations,''
\newblock {\em {IEEE} Comp. Int. Mag.}, vol. 12, no. 3, pp. 18--28, 2017.

\end{thebibliography}

\end{document}